\documentclass[11pt,onecolumn,letterpaper]{article}
\usepackage{acl2016}
\usepackage{times}
\usepackage{epsfig}
\usepackage{graphicx}
\usepackage{textcomp}
\usepackage{amsmath}
\usepackage{amssymb}
\usepackage{gensymb}
\usepackage{caption}
\usepackage{subcaption}
\usepackage{algorithm}
\usepackage{multirow}
\usepackage[noend]{algpseudocode}

\graphicspath{{images/}}

\aclfinalcopy 

\begin{document}

\title{Where is this? \\Video geolocation based on neural network features}

\author{Salvador Medina \qquad Zhuyun Dai \qquad Yingkai Gao\\
Language Technologies Institute\\
Carnegie Mellon University\\
5000 Forbes Ave, Pittsburgh, PA 15213\\
{\tt\small \{salvadom, zhuyund, yingkaig\}@cs.cmu.edu}}

\maketitle

\begin{abstract}
In this work we propose a method that geolocates videos within a delimited widespread area based solely on the frames visual content. Our proposed method tackles video-geolocation through traditional image retrieval techniques considering Google Street View as the reference point. To achieve this goal we use the deep learning features obtained from NetVLAD to represent images to measure the frame-wise similarity. A family of voting-based methods were proposed to aggregate frame-wise geolocation results which boost the video geolocation result. The best aggregation found through our experiments considers both NetVLAD and SIFT similarity, as well as the geolocation density of the most similar results. To test our proposed method, we gathered a new video dataset from Pittsburgh Downtown area to benefit and stimulate more work in this area. Our system achieved a precision of 90\% while geolocating videos within a range of 150 meters or two blocks away from the original position.
\end{abstract}

\section{Introduction}
\label{sec:introduction}
Unlabeled video is readily available from different sources but has been popularly widespread through social media networks. One specific type of shared videos through social media are egocentric, where the video is recorded from a first person point of view as the ones captured during a trip showing the visited places, or from social events such as parades or festivals. Moreover, it has also become popular to share public safety events such as shootings and manifestations. This last type of videos has become of utter importance while gathering evidence for event reconstruction while gathering evidence. The geolocation of videos would aid to locate public safety events such a robbery or an act of vandalism.

A common issue found while using multimedia from media networks is the lack of geotagging information, as this is removed by the user on purpose or eliminated automatically before becoming public to protect the privacy of the users. Taking the fact that social media networks are a rich and a vast multimedia source, it will be useful to come up with a method capable of geolocating the videos based only on the visual content of these digital videos.

Determining the location of a single frame is a problem solved by the geolocalization of an image, which is a well-developed area. One of the most common practices consists on building a reference database in which the image locations are known and searching for the most similar reference images with respect to a query image. The image locations are usually provided in geodetic coordinates: latitude, longitude, and altitude. The set of reference images can be obtained from services such as Google Street View (GSV), which provide multi-angle images taken along major roads, or these could also be obtained from photograph hosting services such as Flickr, since these store images with geolocation based on the user input. In modern days, the data is not a critical factor as the definition of a reliable image descriptor, along with a suitable retrieval strategy. Recently, the NetVLAD image descriptor \textit{au pair} an approximate K-Nearest Neighbors (KNN) has been used to solve the image geolocation problem\cite{arandjelovic2015netvlad,torii2013pitt250k}.

In this work, we focus our efforts to geolocate a video based solely on the visual modality. To achieve this we take advantage of state-of-the-art image geolocation techniques to query the keyframes from a video with an anonymous location. Our proposed method aggregates the results obtained from querying the video keyframes to our reference image database based on the Google Street View Pittsburgh 250k dataset. The aggregation result considers different voting schema including simple voting, weighted-voting, and density-based voting. 

The main contributions of our work are: (1) the definition of the task of geolocating videos based only on the visual aspect, (2) the creation a high quality testing dataset that consists of 50 videos taken around Pittsburgh downtown area, (3) the definition of a technique to visualize the matching area of images using NetVLAD features to facilitate its understanding, and (4) the proposal of a novel reference image aggregation strategy that significantly improves the precision of video geolocation given a strong baseline based on image retrieval.

The ideas presented in this paper are organized as follows: Section~\ref{sec:related_work} gives a detailed background related to this project. The characteristics and properties of NetVLAD are introduced in Section~\ref{sec:feature_analysis}.  Section~\ref{sec:video_processing} and Section~\ref{sec:aggregation} describe our video pre-processing strategy and reference image aggregation strategy respectively. The experiments settings and results can be found in Section~\ref{sec:experiments} and Section~\ref{sec:results}. Finally, in Section~\ref{sec:conclusions} and section~\ref{sec:future_work} we give a conclusion and discussion about the next steps and further ideas that might improve the method.

\begin{figure}[t]
\begin{center}
   \includegraphics[width=1.0\linewidth]{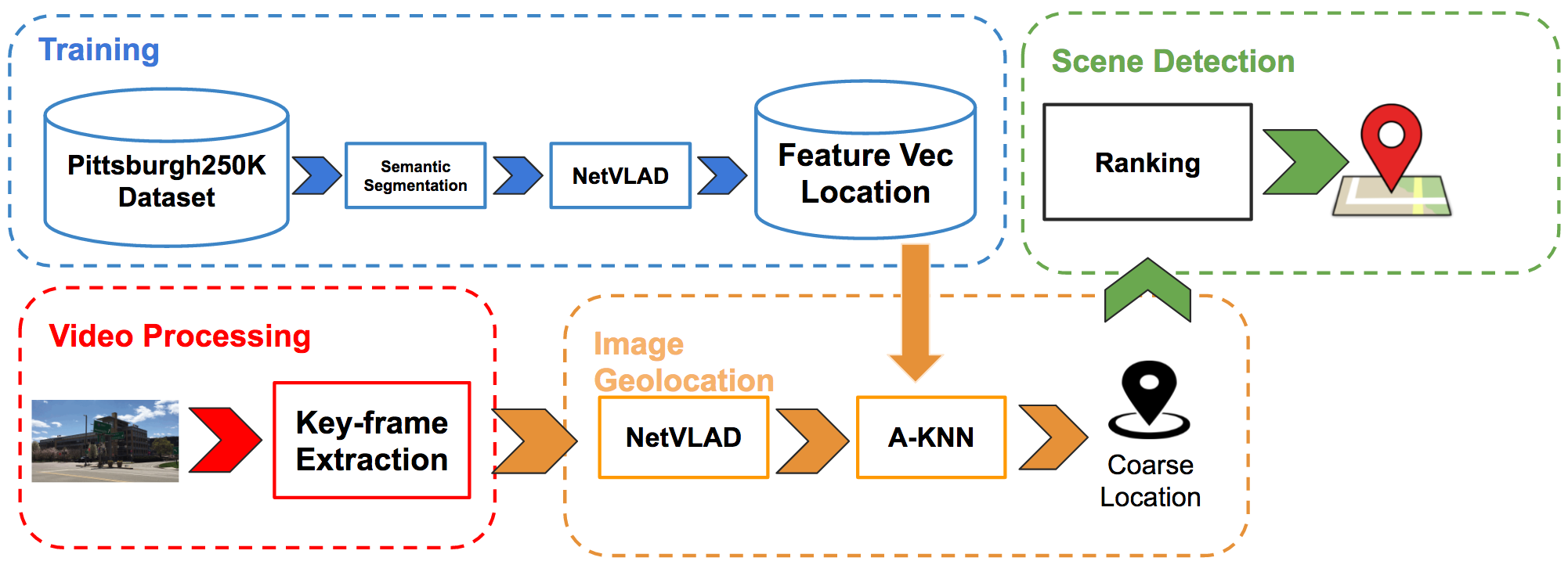}
   
\end{center}
   \caption{System architecture}
\label{fig:system_architecture}
\end{figure}

\section{Related Work}
\label{sec:related_work}
It is to our knowledge that only few research projects address the problem of geolocating videos. Song et al.~\cite{song2012web} proposed a web video geolocating algorithm by propagating geotags among the web video social relationship graph. For this reason, their algorithm can only locate videos from social networks. Snoek et al.~\cite{snoek2011videos} use video content along with various additional signals to infer the location of videos. Their method requires a predefined candidate coarse location set, such as Trafford (UK) and Baghdad (Iraq), where the query video is classified into one of the candidate locations. One limitation of this method is that it restricts the predicted locations to a predefined candidate set. Besides, it only infers the city or town where the video is taken, but cannot give accurate geodetic coordinates. 

In this work we tackle the video geolocation problem through image geolocation techniques, and cast it as an image retrieval task. The problem of image place recognition has been studied extensively \cite{arandjelovic2015netvlad}, \cite{arandjelovic2014dislocation}, \cite{cao2013graph}, \cite{chen2011city}, \cite{gronat2013learning}, \cite{knopp2010avoiding}, \cite{sattler2011fast}, \cite{schindler2007city}. In these proposed methods, usually the query image location is estimated using the locations of the most visually similar images retrieved from a large geotagged database. 

In our approach, we take advantage of NetVLAD features, the current state-of-art image descriptor for geolocation task. The NetVLAD network is formed by replacing the fully connected and soft-max layers from a regular convolutional neural network (CNN) with a NetVLAD layer. This layer takes the convolutional filters from the last layer before the max-pooling phase as input, applies a VLAD \cite{jegou2010aggregating} operation over the convolutions at depth level, to form locally aggregated descriptors which are intra-normalized, concatenated and finally normalized to one with respect to the newly formed vector.

The learning process of NetVLAD is based on the weakly supervised triplet ranking loss which aims to minimize the L2 norm for a learning sample image with respect to its most similar images, while at the same time tries to maximize its distance against dissimilar images. This is the design key component that enables NetVLAD to localize images from a retrieval perspective.

\section{Feature Analysis}
\label{sec:feature_analysis}

As the NetVLAD features are the key component of our proposed method, we performed an exploratory analysis of the features to determine the potential and limitations of the image embedding used in this work.

\subsection{Rotation and Scale Invariance}

\begin{figure}[t!]
\centering
\begin{subfigure}{0.45\textwidth}
    \centering
    \includegraphics[width=1.0\linewidth]{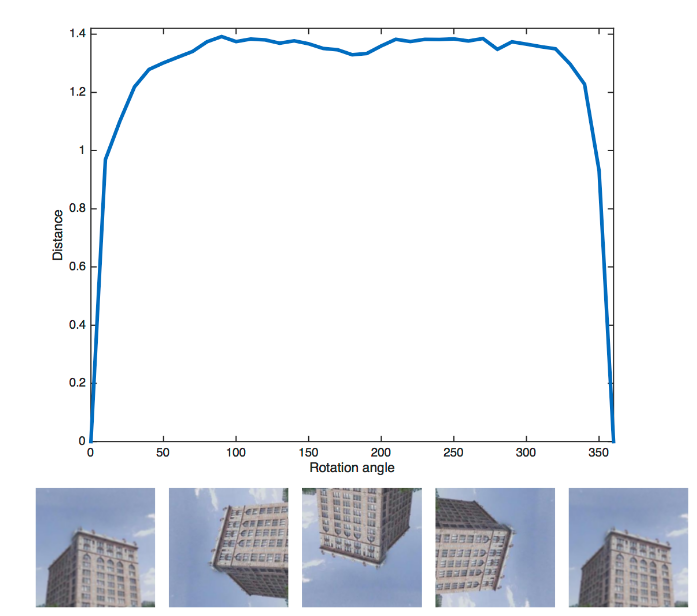}
    \caption{Rotation  Analysis: angle vs distance}
    \label{fig:rotation_invariance}
\end{subfigure}
\begin{subfigure}{0.4\textwidth}
    \centering
    \includegraphics[width=1.0\linewidth]{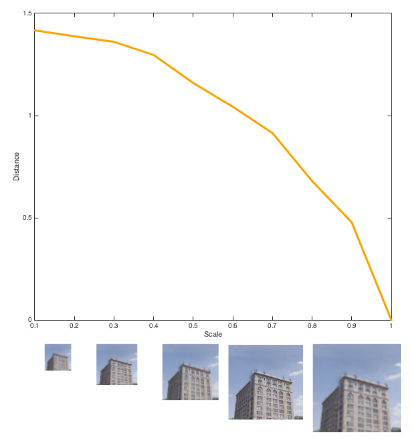}
    \caption{Scale Analysis: scale vs distance}
    \label{fig:scale_invariance}
\end{subfigure}
\label{fig:invariance}
\caption{Invariance Analysis}
\end{figure}

Our first feature analysis aims to determine if the NetVLAD descriptors are invariant to rotation as the features obtained by SIFT \cite{lowe1999object} or AKAZE \cite{alcantarilla2011fast}. In this experiment we fully rotated the image in steps of 10$^{\circ}$. Figure \ref{fig:rotation_invariance} shows the results demonstrating that the NetVLAD image representation is not rotation invariant as the distance increases as soon as we begin to rotate the image. 

Additionally, the scale invariance was tested by scaling down an image by a factor of 0.1 per step from $1$ to a scale of $1/10$. As we can observe in Figure \ref{fig:scale_invariance}, the results of this experiment demonstrate that the NetVLAD feature is not scale invariant, as when we scale down the image from right to left the distance increases which means a decrease in similarity.
 
\subsection{Visualizing the similarity}
\label{sec:visual_attention}

Arandjelovic et al. \cite{arandjelovic2015netvlad} visualize the results of the NetVLAD network by showing a heatmap of the highest activation output value of the CNN to compare against other neural networks, but do not provide a visualization of the neural net's activations based on the actual image resemblance.

In an attempt to visualize the similarity between images by using NetVLAD descriptors, we followed the steps of Zeiler and Fergus \cite{zeiler2014visualizing}, where they partially occlude an image and display in the similarity heatmap the highest classification output value in the location of the partial occlusion. In our case, since the NetVLAD feature is no other than last convolutional layer before max-pooling to which the VLAD operator has been applied, then the image similarity visualization can be achieved through a heatmap which shows the similarity measure between images by comparing the partial occlusion of the querying image against the most similar image from the reference image database.

Our proposed visualization procedure is as follows: First, we calculate the NetVLAD image descriptor of the most similar image from the reference dataset. Then we partially occlude the image with a black patch which must be of the size and location of the convolutional filter from the first layer of the network. Then, we calculate the distance between the occluded querying image with respect to the most similar image and add the similarity value to the block with size and position of the occlusion block into the output heatmap which has the size of the querying image. This procedure is repeated for each of the partially occluded images until the similarity heatmap is completely filled. At the end we normalize the values on the heatmap to be within the range [0,1] and superimpose it to the querying image to visualize the attention of the model.

\begin{figure*}
\centering
\begin{subfigure}{0.33\textwidth}
    \centering
    \includegraphics[width=1.0\linewidth]{./images/netvladviz.png}
    \caption{Procedure}
    \label{fig:visual_attention_proc}
\end{subfigure}
\begin{subfigure}{0.4\textwidth}
    \centering
    \includegraphics[width=1.0\linewidth]{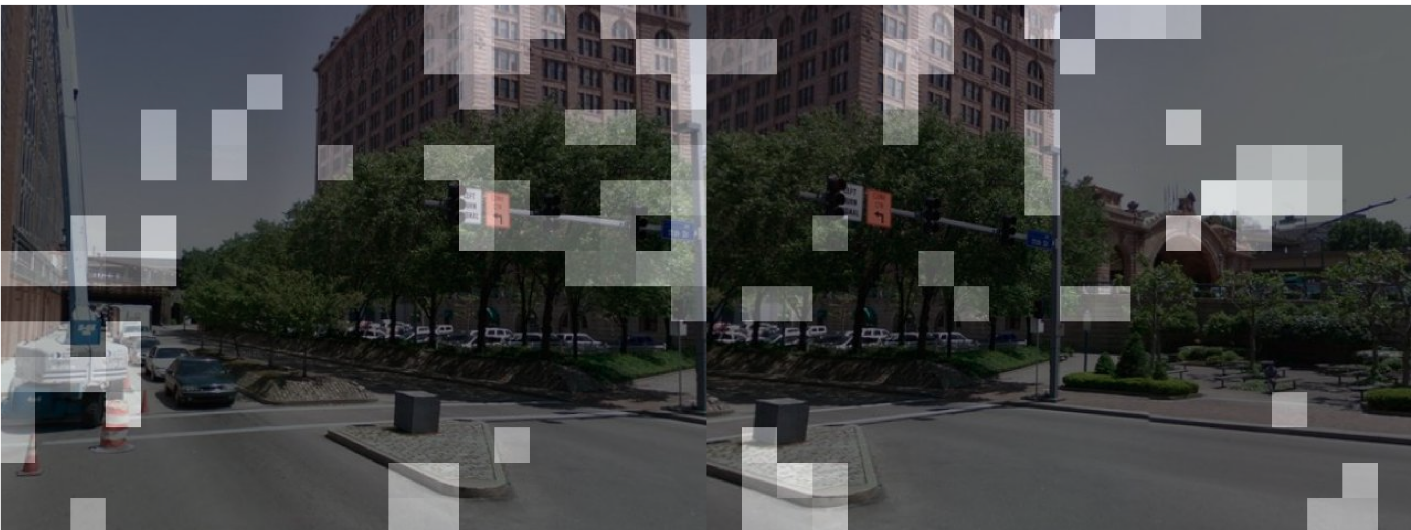}
    \caption{Salient regions example}
    \label{fig:visual_attention_example}
\end{subfigure}
\caption{NetVLAD Visual Attention}
\label{fig:visual_attention}
\end{figure*}

In Figure \ref{fig:visual_attention} we show on the left a graphical description of the procedure aforementioned for its better understanding, and on the right there is an example of the output of our visualization. As we can see in the example the similarity focuses on salient structures as corners of building, street lamps, sidewalks, signs or high contrast of colors between edges. These regions are no different from what other image features such as SIFT or BRIEF \cite{calonder2012brief} focus on for structure. 

\section{Video Processing}
\label{sec:video_processing}
\subsection{Keyframe Extraction}
Before passing to the image retrieval phase, it is necessary to extract the \textit{keyframes} or set of frames that are representative of the query video. Extracting keyframes greatly improves the efficiency of the method as the amount of query images is reduced from the amount of total frames in the video to only a few. The keyframe extraction step also benefits the precision of the method as many of the blurry and noisy frames are filtered out. 

In this work, we use the \textit{difference of color-histograms} for keyframe extraction. The idea behind this method is to select frames that differ in the color-space, as this avoids the selection of redundant scenes. To obtain the keyframes, first we extract the video I-frames, frames that a compressor examines independent of preceding and following frames. Then the color-histogram is calculated for each I-frame by quantizing each of the \textit{RGB} channels into 4 bins, giving a total of $4^3 = 64$ bins. After obtaining the color histograms, we calculate the Manhattan distance of the previously selected keyframe against the next consecutive I-frames' histograms. An I-frame is selected as a keyframe when its distance with respect to the previous keyframe is larger than a certain threshold $T$.



\section{Aggregation Strategy}
\label{sec:aggregation}

In this section we present a family of voting-based methods that aggregates the frame-wise geotagged image retrieval results into one location predicted for the query video. First, we introduce a simple voting method as the baseline. Then we represent weighted-rank voting that bias towards top retrieved images, and density-based voting that favors areas where the candidate location distributions are dense. Finally, we present a weighted voting method that considers the image similarity in both the NetVLAD space and the SIFT space.

\subsection{Simple Voting}
\label{sec: simple_voting}

\emph{Simple voting} is a direct mapping of the original image retrieval problem that Arandjelovic et al. \cite{arandjelovic2015netvlad} formulated for image retrieval. We set it as our baseline.

For simple voting lets consider that each video is formed by $n$ keyframes. The ranking method considers for each keyframe the top $K$ most similar images found in the A-KNN phase of the pipeline. Therefore, each video gets $n \times K$ geotagged images. These images usually overlap in their locations since two or more querying images may be associated with the same location since a localized reference image could be retrieved by different keyframes. One of the main reasons is that GSV is formed by 24 pictures for each location with different angles and yaws, the images from close locations overlap with each other. This redundancy in visual information enables the voting strategy, since the $n \times k$ geotagged reference images are seen as $n \times k$ (or less) \emph{location votes}. The predicted location is the one that has the voting majority.

Nonetheless, using a simple voting scheme conveyed several problems with the reference dataset. For instance, the retrieved $n \times K$ locations contained noisy results or outliers on the map. Another problem occurs when the retrieved locations did not agree and each location got only one vote. In that case, the simple voting method would not perform any better than a random decision.

\subsection{Weighted-Rank Voting}
Due to the problems found in the \textit{simple voting} approach, we explored the \textit{weighted-rank voting} \cite{fishburn1967weightedranking} as a second aggregation method. In \textit{simple voting} we consider the top $K$ most similar images per keyframe to be equally important. However, in reality images that are lower at the retrieval ranking are likely to be less similar to the keyframe, that is, less likely to be at the same location of the video. Instead of giving each image $1$ vote, we assign a different weight to the vote of of each retrieved image as the inverse of its ranking position as described in eq. \ref{eq:weighted_ranking}.

\begin{equation}
vote = \frac{1}{rank}
\label{eq:weighted_ranking}
\end{equation}

\subsection{Density-based Voting}

Weighted-rank voting only takes into consideration the preferences of retrieved images under each keyframe. However, some keyframes can be less informative, or more difficult for the image retrieval process, so the results from these uninformative keyframes are less reliable. As a third method, we considered \textit{density-based voting} which compares all of the candidate locations of a video globally. The intuition is that buildings in the video can appear in multiple geotagged images taken from different places that are close to each other; thus the area where the candidate locations are most concentrated is more likely to cover the correct location of the video. 
Density-based voting considers the top $K$ most similar images per keyframe as previous methods. The main difference is that this method begins by clustering the images based on their geodetic coordinates. Then it only allows the images from the most dense cluster to vote for the location prediction. In this way, the voting is constrained to a smaller area on the map. For this third approach, we selected DBscan \cite{ester1996density} due to its ability to detect outliers.

\begin{figure}[!th]
\centering
\begin{subfigure}[!htb]{0.4\textwidth}
    \includegraphics[width=\textwidth]{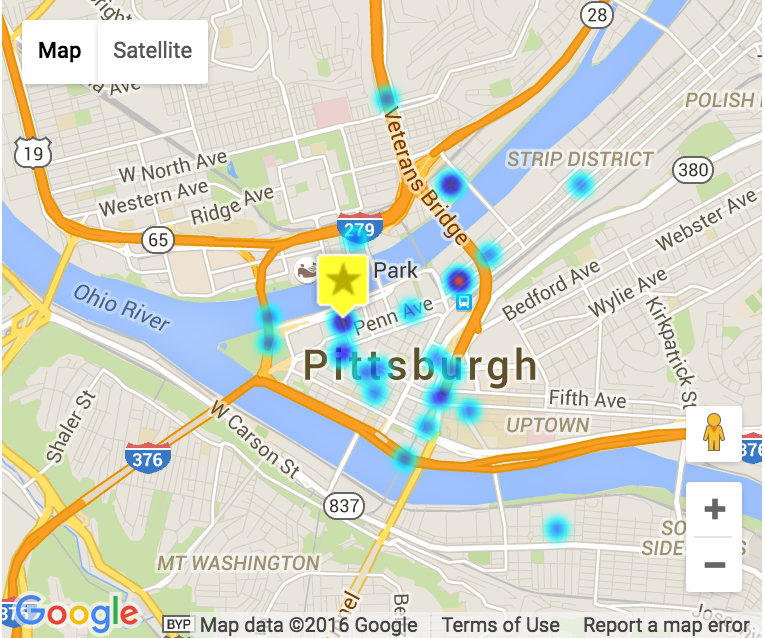}
    \caption{\small Candidate locations shown in Google Maps}
    \label{fig:dbscan_googlemaps}
\end{subfigure}
\begin{subfigure}[!htb]{0.45\textwidth}
    \includegraphics[width=\textwidth]{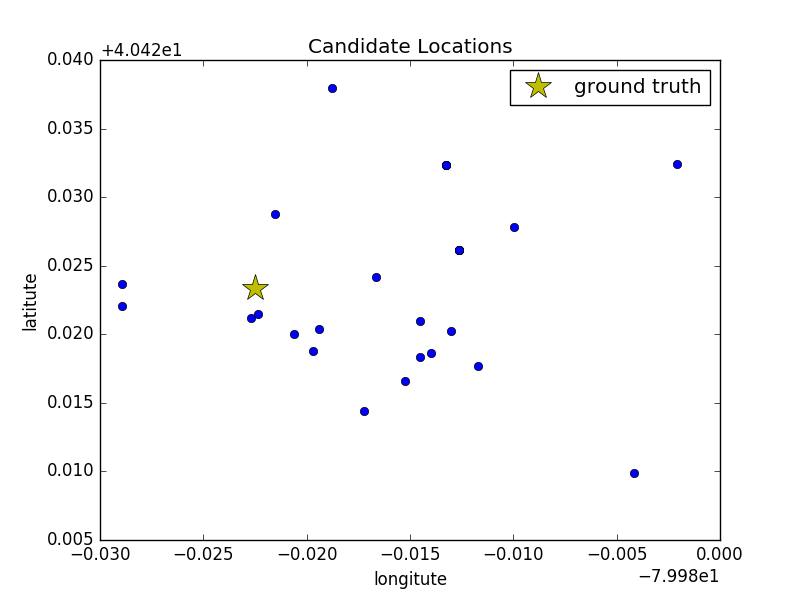}
    \caption{\small Scatter graph of candidate locations}
    \label{fig:dbscan_scattergraph}
\end{subfigure}
\begin{subfigure}[!htb]{0.45\textwidth}
    \includegraphics[width=\textwidth]{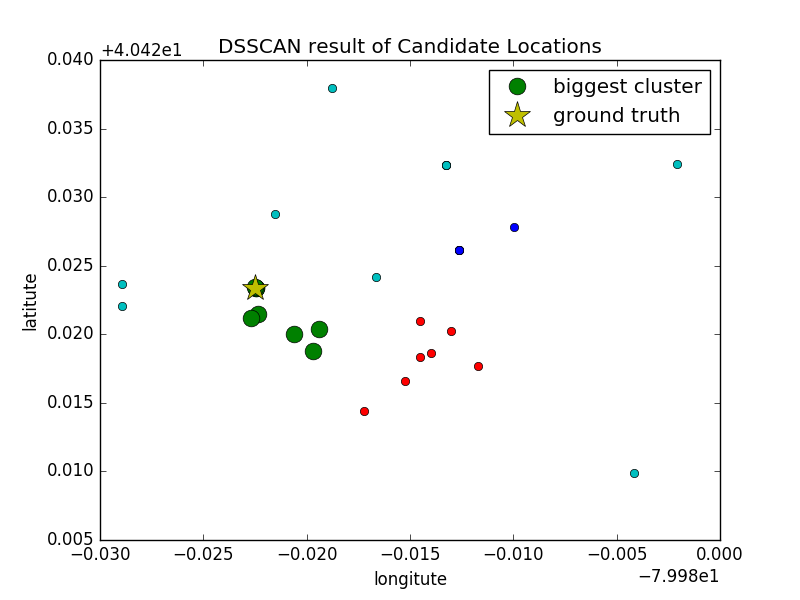}
    \caption{\small Scatter graph of DBSCAN candidate locations}
    \label{fig:dbscan_section}
\end{subfigure}
\caption{\small Comparison of voting methods. }
\label{fig:dbscan}
\end{figure}

Figure \ref{fig:dbscan} gives an example of the results obtained through the proposed methods. Figure \ref{fig:dbscan_googlemaps} shows an example of the location votes heatmap from retrieved geotagged images for a particular video, where the red spots represent more votes on that location, blue represents fewer votes, and the yellow star represents the ground truth. On the other hand, in Figure \ref{fig:dbscan_scattergraph} we find the plot of the coordinates of the candidate locations. This plot reveals why simple voting fails: although many candidate locations are scattered around the ground truth, they do not share exactly the same location and demonstrates that averaging the locations would lead to an error. In contrast, Figure \ref{fig:dbscan_section} shows the result after clustering locations based on DBSCAN, where each color represents a cluster: the light blue dots are outliers, while the thick green markers represent the cluster with the largest population. In our method, only the cluster which has the largest density is allowed to vote. As we can see, by adding the clustering step into the method filters out most of the false-positive locations as the ground truth is usually located within the cluster that has the highest density. 

\subsection{NetVLAD + SIFT}

As previously described, we use NetVLAD features to retrieve geotagged images. Thus the weighted-rank voting method are also based on the similarity of NetVLAD features. In our analysis, although NetVLAD had good precision in general, it did make errors in some cases. The next method considers another widely used visual feature, SIFT, to refine the voting. Within our method we assigned a \textit{SIFT score} to the images which is the average of the distances among the top matching keypoints that were obtained through a brute force approach between the querying keyframe and the top $S$ ranked images obtained through the NetVLAD feature matching.

As in our previous methods, we retrieve the top $K$ most similar geotagged images per keyframe based on the NetVLAD features. Then we extract SIFT features for these images and keyframes. Each geotagged image is assigned with a weight that linearly combines the NetVLAD ranking and the SIFT similarity score as described in equation \ref{eq:netvlad_sift}.

\begin{equation}
vote = \lambda \frac{1}{rank_{netvlad}} + (1-\lambda) score_{SIFT}
\label{eq:netvlad_sift}
\end{equation}

\section{Experiments}
\label{sec:experiments}

\subsection{Datasets}
\paragraph{Pittsburgh 250K} The dataset used for building the geotagged reference dataset was the Pittsburgh 250K which is formed by 254,064 images obtained from GSV. Each image has a resolution of 640x480 pixels and were taken at 10,586 different locations mainly in Pittsburgh Downtown. At each location the images were captured with 12 yaws of 30$^{\circ}$ and two pitches one at floor level with 0$^{\circ}$ and the second with a 30$^{\circ}$ angle above the ground.

\paragraph{Pittsburgh Urban Video} For testing purposes we built a testing video dataset named \textit{Pittsburgh Urban Video}. This video corpus is formed by 50 geotagged videos that were recorded at a resolution of $1920\times1080$ pixels. Most of the videos simulate a tourist recording while visiting a city. The majority of the videos were recorded by panning urban views which also include a high tilt into skyscrapers found in the Pittsburgh downtown area. We incorporated recordings in landscape and portrait mode, rotated videos, fast movement, quick zoom-in and zoom-out and close-ups to objects which might not provide great information as waste bins, flower pots or traffic lights. By creating a custom test set we could ensure that the test images were not included in the training set and validate the performance of the proposed method.

\subsection{Experimental Setup}

\textbf{NetVLAD feature extraction}. The geolocation system required to compute offline the NetVLAD descriptors from all the images in the Pittsburgh 250K dataset and all the keyframes from the Pittsburgh Urban Video dataset. Due to the large scale of the processing, we distributed the workload between 25 different computing nodes, in batches of approximately 10,000 images. After each node finished its task, the data was gathered into one single data block. The distribution of the workload reduced the computational time from an estimate of $\approx625$ hours down to $\approx25$ hours.

Once the descriptors from the training and testing dataset were obtained, we performed a series of experiments to determine which method is the most suitable for geolocalizing videos. 

\textbf{Evaluation}. Location coordinates can be considered as a continuous space, and this makes it nearly impossible to measure the system's precision or recall based on an exact location prediction. For this reason, we evaluate our methods through the precision of the predicted location within a certain distance range $d$ as described in eq. \ref{eq:location_prediction}.

\begin{equation}
P(d) = \frac{\sum_{v=1}^{|V|}1\{dist(pred_v, truth_v) <= d\}}{|V|}
\label{eq:location_prediction}
\end{equation}

Where $|V|$ is the total number of testing videos, $dist(pred_v, truth_v)$ is the spatial distance from the predicted location to the ground truth. A prediction is considered to be correct if it is within $d$ meters from the ground truth. However, as $d$ goes larger, even randomly picking locations in the city could achieve $100\%$ precision, and the measure becomes meaningless. In this work, we consider the prediction to be meaningful only if it is within a block from the ground truth. As a coarse estimation, the average block size in the Pittsburgh Downtown area is around $65m \times 150m$ ($0.04mi \times 0.1mi$). Therefore the final evaluation measures precision in the range of $d \in [0,150]$ meters.

\section{Experimental Results}
\label{sec:results}
In this section, first we compare the voting scheme to a random baseline, then we evaluate the video geolocation precision for each of the aggregation methods, and finally compare them to an oracle system. Finally, we show a comparison of the performance of all the aggregated methods.




\subsection{Video Geolocation Random Baseline}
Voting methods assume the existence of an overlap from the retrieved results. As if the different possible predictions have only one vote, the voting would be no different from a random decision. To confirm the overalpping assumption, our first experiment compares the simple voting method against a random baseline.

The random baseline was built by considering the top $K$ most similar images per keyframe as candidates, which results in $n \times K$ candidate geotagged images for an $n$ keyframes video. Then the random baseline selects the location of one randomly selected image out of the $n \times K$ reference images as the prediction of the system. 

\begin{figure*}[!th]
\begin{subfigure}{.6\textwidth}
  \centering\captionsetup{width=1.0\linewidth}%
  \includegraphics[width=1.0\linewidth]{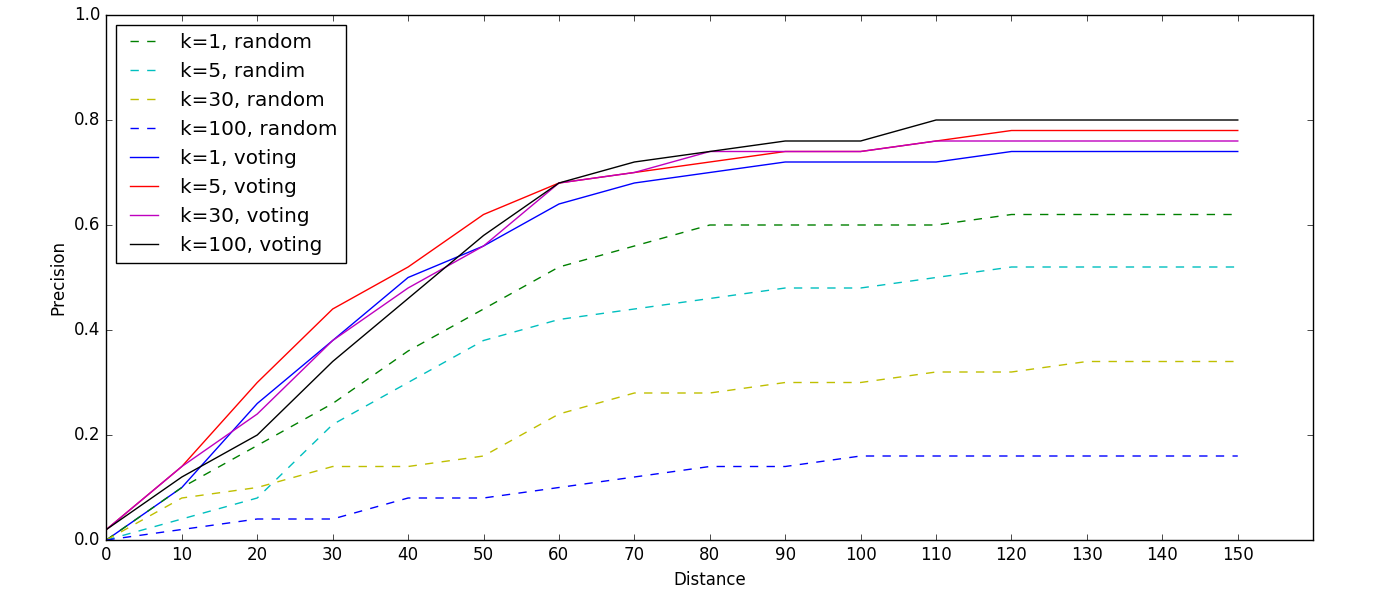}
  \caption{\small Comparison of random baseline (dashed line) and simple voting (solid line) at top $K$ candidate reference images per keyframe. X-axis: distance from ground truth, Y-axis: video geolocation precision from eq.\ref{eq:location_prediction}.}
  \label{fig:random}
\end{subfigure}%
\begin{subfigure}{.4\textwidth}
  \centering\captionsetup{width=.8\linewidth}%
  \includegraphics[width=1.0\linewidth]{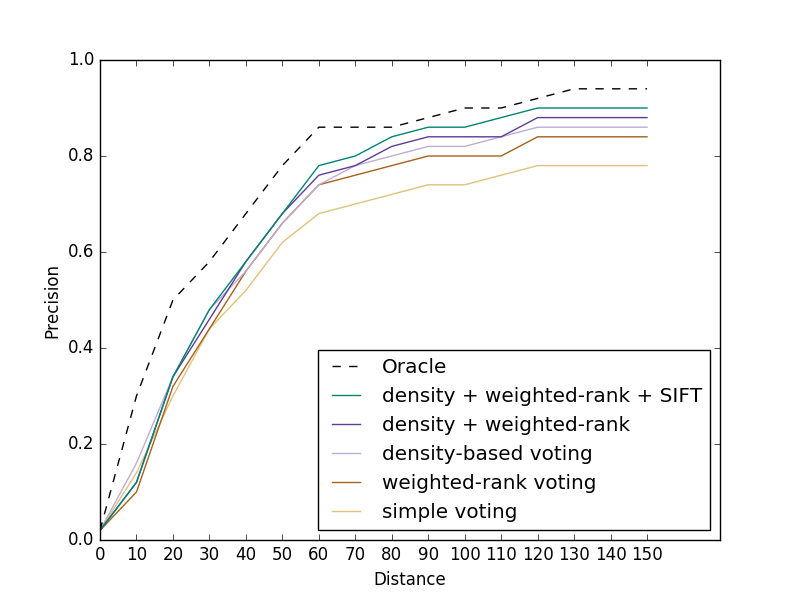}
  \caption{\small Comparison of proposed aggregation methods. X-axis is the distance from ground truth, Y-axis is the video geolocation precision as defined in eq.\ref{eq:location_prediction}.}
  \label{fig:res_all}
\end{subfigure}%
\label{fig:results}
\end{figure*}

Figure \ref{fig:random} shows the performance of the random baseline and the simple voting, at various levels of $K$. The random baseline is relatively strong when $K=1$, as the random choices are narrowed down to the most similar images. Nonetheless, the voting method performance is better than the best random performance. This experiment proves the assumption of location overlap. It demonstrates that voting can aggregate evidence from retrieved images and improve video location prediction. The location overlap can clearly be seen in Figure \ref{fig:retrieval_results} where the most salient results tend to be from nearby locations rather from the same location due to the nature of the salient similar structures such as buildings, bridges or even billboards.

\begin{figure}[t!]
\begin{center}
\end{center}
   \caption{Retrieval result example}
\label{fig:retrieval_results}
\end{figure}

\subsection{Video Geolocation Effectiveness}
The next experiment analyzes the performance of the proposed aggregation methods. We also tested an \emph{oracle system}: consider $K$ candidates per keyframe and select the candidate location that is closest to the ground truth as the prediction. This is the upper-bound of any aggregation method given a fixed image retrieval output. The performance of the oracle system also reflects the effectiveness of the NetVLAD image retrieval process.

Figure \ref{fig:res_all} reports the video geolocation performance of 5 proposed methods: simple voting, weighted-rank voting, density-based voting with equal weight, density-based voting with weighted-rank, and density-based voting with NetVLAD+SIFT weighting. The oracle performance is in dashed black line. All methods considered $K=5$. For NetVLAD+SIFT weighting scheme, we selected $\lambda=0.4$ to be the best value after a coarse parameter sweep.
As shown in Figure \ref{fig:res_all}, the precision gradually improved as more methods were aggregated. The best method considers voting, location density, NetVLAD similarity and SIFT similarity. It is worth noticing that in short error distance all the methods are much lower than the oracle. This means that some near-correct locations are retrieved from the database, but our aggregation methods fails to select them as the final prediction. As the error distance threshold increase, the differences between each method also increases, pushing the best method closer to the oracle precision.

As shown in Table \ref{tab:res}, the NetVLAD+SIFT+Density improved the precision over the simple voting by over $10\%$. $90\%$ of its predictions are within 150 meters or two blocks from the original location. Moreover, at the 150-meter distance, NetVLAD+SIFT+Density is only $4\%$ worse than the oracle.

\begin{table*}[!th]
\small
\centering
\caption{Precision of the random baseline, the simple voting baseline the best voting method, NetVLAD + SIFT + Density, and the oracle system. Parameters: $K=1$ for Random, $K=5$ for Simple voting, $K=5, \lambda=0.4$ for NetVLAD+SIFT+Density, and $K=5$ for the oracle system. Percentage in parenthesis shows relative gains/loss from simple voting.}
\label{tab:res}
\begin{tabular}{c|l|l|l|l|l|l}
\hline \hline
\multirow{2}{*}{Method} & \multicolumn{6}{c}{Distance ($d$, in meters)}        \\ \cline{2-7} 
                        & 5     & 10    & 30    & 50    & 100   & 150   \\ \hline
Random                  & 0.00  & 0.04  & 0.16  & 0.28  & 0.36  & 0.42  \\ 
Simple Voting           & 0.04  & 0.14  & 0.44  & 0.62  & 0.74  & 0.78  \\
\textbf{NetVLAD+SIFT+Density} & \textbf{0.06 (+ 50\%)} & 0.12 (- 14\%) & \textbf{0.48 (+ 9\%)} & \textbf{0.68 (+ 10\%)} & \textbf{0.86 (+16\%)} & \textbf{0.90 (+15\%)} \\ 
Oracle                  & 0.1   & 0.3   & 0.58  & 0.78  & 0.9   & 0.94  \\ \hline \hline
\end{tabular}
\end{table*}

\section{Discussion}
\label{sec:conclusions}

In this work, we have presented the results of a powerful method capable of localizing videos by combining traditional retrieval techniques such as mixed ranking along with new neural net based image features. During our experiments, the NetVLAD feature space showed a good performance for a relatively large scale image corpus, however it was shown that these features are not invariant to scale or rotation, which limit the model and potentially perform poorly on videos taken with a tilted camera.

We also presented a method to visualize the attention of the NetVLAD model while looking at the similarity heatmap between two images. The method proved to be insightful to understand the attention of the model to structures and colors within a city landscape.

In our experiments we showed that a traditional information retrieval approach as aggregating ranking methods for image retrieval improves the geolocation of videos. We found that the best model in our experiments was the mixed voting of weighted and density based ranking that incorporates SIFT matching scores. The final mixed voting method showed a precision of 0.9 at a distance of 150 meters which are approximately two blocks within Pittsburgh Downtown.

\section{Future Work}
\label{sec:future_work}

The analysis based on the oracle system showed that the aggregation fails to fully select the correct near locations, which indicates that there is still room for improvement to our method based on the Pittsburgh Urban Video dataset.

After observing the images in the Google Street View Pittsburgh 250k dataset, we realized that one possible path which might increase the performance of our method is to occlude the sky while measuring the similarity between images. Shen and Wang \cite{shen2013sky} proposed a method to segment the sky by an incremental detection of the horizon which is suitable to the type of images found in urban GSV images. Another approach is to use neural network based image segmentation methods such as Mask R-CNN \cite{he2017mask} or DeepLab \cite{chen2018deeplab}.

It also came to our attention that in our best method the SIFT matching score provided a larger weight into the final vote for the location of the videos. This is an invitation to explore the incorporation of more traditional image features such as AKAZE~\cite{alcantarilla2011fast} or BRIEF~\cite{calonder2012brief}, where the latter might be the most suitable descriptors due to their low computational cost given the large scale of the problem.

As a final note, since  NetVLAD features lack rotation and scale invariance, it is also recommended to augment the reference database by randomly rotating and scaling the original content to overcome this limitation.


\end{document}